\documentclass[11pt]{article}

\usepackage[preprint]{acl}

\usepackage{graphicx}
\usepackage{subcaption} 
\usepackage{multirow}
\usepackage{times}
\usepackage{latexsym}
\usepackage{amsmath}
\usepackage{enumitem}
\usepackage{caption}
\usepackage[T1]{fontenc}
\usepackage{algorithm}
\usepackage[utf8]{inputenc}
\usepackage{amssymb}
\usepackage{algpseudocode}
\usepackage{booktabs} 
\algrenewcommand\algorithmicrequire{\textbf{Input:}}
\algrenewcommand\algorithmicensure{\textbf{Output:}}
\usepackage{microtype}
\usepackage{bm}
\usepackage{inconsolata}
\usepackage[table]{xcolor} 

\usepackage{graphicx}

\usepackage{color, xspace}

\usepackage[most]{tcolorbox}

\newtcolorbox{promptbox}[2][blue]{%
  enhanced,
  breakable,
  colback=#1!4!white,
  colframe=#1!60!black,
  boxrule=0.5pt,
  arc=2pt,
  left=4pt,right=4pt,top=4pt,bottom=4pt,
  title={#2},
  fonttitle=\bfseries
}

\title{How to Utilize Complementary Vision-Text Information for 2D Structure Understanding}

\author{
\textbf{Jiancheng Dong}$^{\mathsection\dagger}$, 
\textbf{Pengyue Jia}$^{\mathsection}$, 
\textbf{Derong Xu}$^{\mathsection}$, 
\textbf{Jiawei Cheng}$^{\mathsection}$, 
\textbf{Jingyu Peng}$^{\mathsection\dagger}$,
\textbf{Chao Zhang}$^{\mathsection}$, \\
\textbf{Bowen Liu}$^{\mathsection}$, 
\textbf{Xin Sun}$^{\dagger}$, 
\textbf{Lixin Su}$^{\dagger}$, 
\textbf{Shuaiqiang Wang}$^{\dagger}$, 
\textbf{Dawei Yin}$^{\dagger}$, 
\textbf{Xiangyu Zhao}$^{\mathsection} \footnotemark[1]$
  \\
  $^{\mathsection}$ City University of Hong Kong, $^{\dagger}$ Baidu Inc. \\
  \texttt{jiancdong2-c@my.cityu.edu.hk}
}

\begin{document}
\maketitle
\begin{abstract}
LLMs typically linearize 2D tables into 1D sequences to fit their autoregressive architecture, which weakens row-column adjacency and other layout cues. In contrast, purely visual encoders can capture spatial cues, yet often struggle to preserve exact cell text. Our analysis reveals that these two modalities provide highly distinct information to LLMs and exhibit strong complementarity.
However, direct concatenation and other fusion methods yield limited gains and frequently introduce cross-modal interference. To address this issue, we propose DiVA-Former, a lightweight architecture designed to effectively integrate vision and text information. DiVA-Former leverages visual tokens as dynamic queries to distill long textual sequences into digest vectors, thereby effectively exploiting complementary vision--text information.
Evaluated across 13 table benchmarks, DiVA-Former improves upon the pure-text baseline by 23.9\% and achieves consistent gains over existing baselines using visual inputs, textual inputs, or a combination of both.
\end{abstract}

\begin{figure}[t]
    \centering
    \makebox[\columnwidth][r]{%
        \begin{minipage}{\columnwidth}
            \centering
            \begin{subfigure}{\linewidth}
                \centering
                \includegraphics[width=0.95\linewidth]{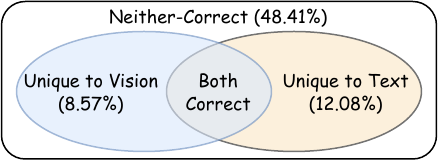}
                \caption{Complementarity between vision and text.}
                \label{fig:fig1a}
            \end{subfigure}

            \vspace{0.5em}

            \begin{subfigure}{\linewidth}
                \centering
                \includegraphics[width=\linewidth]{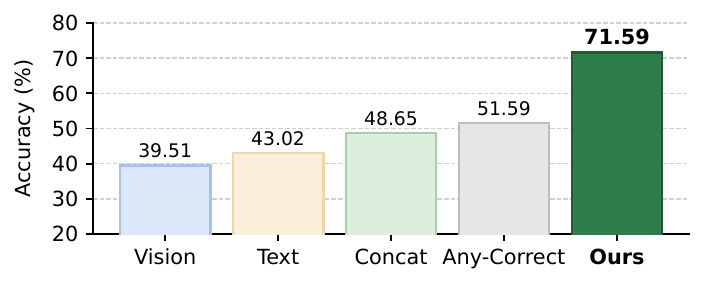}
                \caption{Accuracy comparison of different methods.}
                \label{fig:fig1b}
            \end{subfigure}
        \end{minipage}%
    }
    \caption{\textbf{Pilot study on TCR.} “Any-Correct” denotes the proportion of examples for which either the vision or text modality yields the correct answer.
}
    \label{fig:intro}
\end{figure}

\section{Introduction}
\label{sec:Introduction}

Large Language Models (LLMs) have recently shown strong capability in table understanding \cite{liu2024rethinking, zhao2023large, he2025tablelora}. However, most existing LLM-based methods still rely on serializing 2D tables into 1D token sequences so that the inputs fit the sequence modeling paradigm of LLMs \cite{yang2022tableformer}. Such a 2D-to-1D flattening process inevitably breaks native row-column adjacency and weakens access to spatial and structural cues \cite{li20252d,liu2025structured}. In contrast, purely vision-based approaches process table images more naturally in the 2D space, but often struggle to preserve fine-grained textual content, symbols, and exact cell semantics \cite{zheng2024multimodal, chen2023tablevlm, zhang2024unitabnet}. This motivates a basic question: \emph{do vision and text provide complementary evidence for 2D table understanding?}

To answer this question, we first conduct a pilot study on the Table Cell Retrieval (TCR) task. Given an input table and a target cell content, TCR requires the model to predict the row and column indices of the corresponding cell. We choose this task because it largely avoids the interference of language generation priors and directly reflects how much structural information an LLM can capture from a table. Our results reveal a clear discrepancy between the coordinates correctly localized by text-only and vision-only models. As shown in Figure~\ref{fig:intro}(a), among all coordinates successfully localized by either unimodal model, around 40\% can only be solved by one modality but not the other. This finding suggests that visual and textual inputs provide substantially different evidence for 2D structure understanding, and that LLMs solve different subsets of instances depending on the input modality. In other words, the two modalities are highly complementary.

A natural next step is to combine both modalities by directly concatenating visual and textual contexts before feeding them into the LLM. As illustrated in Figure 1(b), direct concatenation improves over unimodal baselines on TCR, but the gains remain limited. It introduces long and redundant context, and leaves cross-modal interaction entirely to the frozen LLM’s self-attention. As a result, it remains ineffective in many cases that cannot be solved by either unimodal model alone. This raises a practical follow-up question: \emph{how can we efficiently utilize the complementary vision-text information for 2D structure understanding?}

To address this question, we propose \textbf{DiVA}-Former (\textbf{Di}gesting with \textbf{V}isual \textbf{A}nchors), a novel and lightweight fusion architecture designed to genuinely exploit complementary vision-text information. DiVA-Former is a compact resampling module with only two Transformer layers. It uses visual tokens as dynamically initialized queries to attend over textual features, compressing the lengthy textual context into a compact sequence of structurally informed \textit{digest vectors}. In this way, visual signals serve as anchors that guide the model in selecting and aggregating the most relevant textual evidence, enabling deeper cross-modal interaction. As shown in Figure~\ref{fig:intro}(b), our method substantially improves coordinate localization accuracy, significantly outperforming direct concatenation and demonstrating a much stronger ability to recover cases that neither unimodal model can solve. These results suggest that DiVA-Former enables more effective cross-modal alignment for 2D structure understanding.

Our contributions can be summarized as follows:
\begin{itemize}[leftmargin=*] 
    \item We conduct a pilot study on the TCR task and demonstrate that visual and textual modalities provide highly complementary information for 2D structure understanding in tables.
    \item We propose \textbf{DiVA-Former}, a lightweight multimodal fusion framework that effectively exploits such complementarity to improve cross-modal alignment for table understanding.
    \item Extensive experiments across 13 mainstream table benchmarks demonstrate the effectiveness of our method. Under identical data settings, DiVA-Former outperforms the pure text baseline by 23.9\%, the direct concatenation approach by 22.5\%, and adapter-based or resampler-based methods by 8.6\% to 35.8\%.
\end{itemize}
\begin{figure*}[t]
\centering
\includegraphics[width=\linewidth]{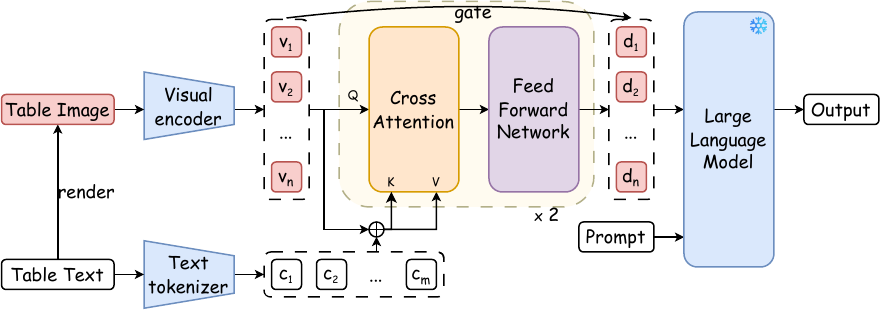}
\caption{The overall architecture of DiVA-Former. The model leverages spatial visual tokens ($\boldsymbol{v}_i$) extracted from a table image as dynamic queries. These queries attend to the linearized table text tokens ($\boldsymbol{c}_j$) via Cross-Attention to generate compact digest vectors ($\boldsymbol{d}_i$). A gating mechanism maintains an approximate identity mapping from the initial visual tokens to the output digest, preserving high-quality 2D spatial priors. 
}
\label{fig:method}
\end{figure*}

\section{Related Work}
\label{sec:related}

\paragraph{Multimodal Table Understanding}
Multimodal table understanding has recently emerged as a rapidly developing field \cite{zhang2025exploring, chen2023tablevlm, zhang2024unitabnet, xu2021layoutlmv2, zheng2024multimodal, mathur2024knowledge, gautam2025tabcomp}. To approach the ``Any-Correct'' upper bound (Figure~\ref{fig:intro}(b)), some recent works \cite{deng2024tables, zhou2025texts} employ rule-based routing to select between visual and textual inputs. However, such methods rely on modality selection rather than true multimodal fusion, and their evaluations are largely limited to TableQA settings. Several studies \cite{zhang2023crt, bhandari2024exploring, zhou2024freb} have demonstrated that LLMs often bypass genuine 2D structural reasoning on TableQA tasks, instead relying on strong linguistic priors to guess the answers. In contrast, we evaluate our method across 13 tabular datasets covering four task categories. By genuinely integrating both modalities, DiVA-Former surpasses the oracle routing upper bound, successfully resolving instances where unimodal baselines fail.

\paragraph{Digesting Context into Soft Embeddings}
Transforming sequential text into informative and compact soft embeddings using purely textual approaches has been an active area of research \cite{yen2024long, dai-etal-2025-pretraining, chevalier-etal-2023-adapting, ge2023incontext, dong2025behavior, kuratov2025cramming, wang2024context, chen2024squid, liu2025context}. Since the introduction of DeepSeek-OCR \cite{wei2025deepseek, wei2026deepseekocr2visualcausal}, the paradigm of using visual tokens to read and encode textual information has gained increasing attention \cite{li2025text, cheng2025glyphscalingcontextwindows, xing2025vision, xing2025see}. 
However, existing soft embedding methods struggle to surpass uncompressed baselines due to inherent information loss (see Appendix~\ref{app:soft} for a detailed empirical analysis). Meanwhile, current "visual text compression" strategies frequently rely on language model priors rather than genuinely interpreting the image text \cite{liang2026visualmeritlinguisticcrutch}. These unaddressed gaps directly motivate our pilot study on the TCR task. By exploiting 2D spatial information from visual tokens, DiVA-Former overcomes these bottlenecks and outperforms uncompressed baselines in table tasks.

\section{Method}
\label{sec:method}

\subsection{Task Formulation}
Given a table understanding sample, we consider three inputs: the serialized table text, the corresponding table image, and a user instruction prompt. Specifically, the serialized 1D table text is tokenized and passed through the embedding layer of a frozen large language model (LLM), producing textual embeddings,
$\boldsymbol{C} = [\boldsymbol{c}_1, \dots, \boldsymbol{c}_m] \in \mathbb{R}^{m \times d}$.
Meanwhile, the table image is processed by a frozen visual encoder to extract 2D-aware visual embeddings,
$\boldsymbol{V} = [\boldsymbol{v}_1, \dots, \boldsymbol{v}_n] \in \mathbb{R}^{n \times d}$.
We denote the user instruction as $p$ and the expected target sequence as $y$.
Our objective is to enable the LLM to exploit the complementary information from both textual and visual modalities for more accurate generation under the same instruction $p$. A straightforward solution is to directly concatenate the visual embeddings $\boldsymbol{V}$ with the textual embeddings $\boldsymbol{C}$ and feed the combined sequence into the LLM as the input context. 

\subsection{DiVA-Former Architecture}
We propose \textbf{DiVA-Former}, a lightweight query resampler that fuses dual-modal information into a compact sequence of digest vectors,
$\boldsymbol{D} = [\boldsymbol{d}_1, \dots, \boldsymbol{d}_n]$.
As illustrated in Figure~\ref{fig:method}, DiVA-Former takes textual embeddings $\boldsymbol{C}$ and visual embeddings $\boldsymbol{V}$ as input, and progressively transforms the visual tokens into multimodal digest vectors that can be consumed by the frozen LLM.

\paragraph{Visual-Token Initialized Dynamic Queries.}
Unlike prior query resamplers that employ a fixed set of learnable queries initialized from scratch~\cite{wang2024context, li2023blip, NEURIPS2022_960a172b}, DiVA-Former derives its queries directly from the visual embeddings. Formally, the initial query states are defined as $\boldsymbol{Q}^{0} = \boldsymbol{V}$.
This design makes both the query length and query initialization adaptive to the visual input. Since the resulting digest sequence preserves the same length as the visual tokens, the representation is typically much shorter than the original textual context. In practice, the number of visual tokens is often around half of the text length, which substantially reduces the computational burden on the frozen LLM.

\paragraph{Cross-Attention Resampler.}
DiVA-Former consists of $N$ stacked transformer layers. At layer $\ell$, the model updates the query states $\boldsymbol{Q}^{\ell-1}$ using a Pre-Norm Cross-Attention block followed by a standard Feed-Forward Network (FFN)~\cite{pmlr-v139-jaegle21a}. The query input comes from the previous-layer digest representation, while the key-value memory sequence is formed by concatenating the text context with the current query states:
\begin{equation}
\boldsymbol{M}^{\ell} = [\boldsymbol{C}; \boldsymbol{Q}^{\ell-1}].
\end{equation}
The cross-attention operation allows each query token to attend not only to the textual embeddings but also to the current digest states themselves, thereby supporting both cross-modal fusion and iterative refinement across layers.

\paragraph{Identity-Preserving Gating.}
To preserve the original visual information and avoid catastrophic forgetting during early training~\cite{NEURIPS2022_960a172b, 10508743}, we introduce an identity-preserving gating mechanism. Each transformer block contains two learnable scalar gates, $g_{\text{attn}}$ and $g_{\text{ffn}}$, both initialized to a near-zero constant $\epsilon$ (denoted as the initial gate value $g_0$). The residual updates at each layer are defined as
\begin{equation}
\boldsymbol{Q} \leftarrow \boldsymbol{Q} + g_{\text{attn}} \cdot \mathrm{CrossAttn}(\boldsymbol{Q}, \boldsymbol{M}, \boldsymbol{M}).
\end{equation}
\begin{equation}
\boldsymbol{Q} \leftarrow \boldsymbol{Q} + g_{\text{ffn}} \cdot \mathrm{FFN}(\boldsymbol{Q}).
\end{equation}
With near-zero initialization, the attention and FFN branches contribute only minimally at the beginning of training, making the block behave approximately like an identity mapping ($\boldsymbol{Q} \approx \boldsymbol{Q}^{0}$). This near-identity residual path stabilizes optimization and enables textual information to be incorporated gradually into the visual queries. As a result, DiVA-Former can be trained directly through instruction fine-tuning without requiring complicated multi-stage pre-training objectives~\cite{ren-etal-2024-learning}.

\subsection{Optimization}
Let $\boldsymbol{\Phi}$ denote the target LLM, and let $R_{\theta}$ denote DiVA-Former parameterized by $\boldsymbol{\theta}$. Given the textual embeddings $\boldsymbol{C}$ and visual embeddings $\boldsymbol{V}$, the resampler produces the final digest \begin{equation}
\boldsymbol{D} = R_{\theta}(\boldsymbol{C}, \boldsymbol{V}) = \boldsymbol{Q}^{N} = [\boldsymbol{d}_1, \dots, \boldsymbol{d}_n].
\end{equation}
During instruction fine-tuning, the compact digest vectors $\boldsymbol{D}$ are used in place of the original verbose table context. We prepend $\boldsymbol{D}$ to the user prompt $p$ and feed the resulting sequence into the frozen LLM $\Phi$. The training objective is the standard teacher-forcing negative log-likelihood over the target sequence $y$:
\begin{equation}
\mathcal{L}_{\text{SFT}} = \mathbb{E}\left[- \sum_{t} \log P_{\Phi}\left(y_t \mid \boldsymbol{D}, p, y_{<t}\right)\right].
\end{equation}
During optimization, gradients are backpropagated only through the lightweight parameters of $R_{\theta}$, including the scalar gates, while all parameters of the target LLM $\boldsymbol{\Phi}$ remain strictly frozen.

\begin{table*}[ht]
\centering
\resizebox{\textwidth}{!}{
\begin{tabular}{llcccccccccccccc}
\toprule
\multirow{2}{*}{Method} & \multirow{2}{*}{Modality}
& \multicolumn{5}{c}{Question Answering}
& \multicolumn{4}{c}{Structure Understanding}
& \multicolumn{2}{c}{Fact Verification}
& \multicolumn{2}{c}{Text Generation}
& \multirow{2}{*}{Avg} \\
\cmidrule(lr){3-7}\cmidrule(lr){8-11}\cmidrule(lr){12-13}\cmidrule(lr){14-15}
& & WTQ & FeTaQA & TAT & HiTab & TabMWP
& TSR & TCE & RCE & TCR
& TabFact & InfoTabS
& RotoWire & WikiBio & \\
\midrule
\multirow{3}{*}{Direct} & Vision & 55.1 & 17.4 & \underline{75.8} & 63.5 & 74.1 & 14.8 & 36.1 & 14.2 & 39.5 & 72.0 & \textbf{77.6} & 13.9 & 13.1 & 43.6 \\
 & Text & 54.6 & 19.2 & 73.9 & \underline{68.4} & 75.9 & 14.2 & 24.2 & 13.8 & 43.0 & 72.4 & 72.8 & 15.2 & 13.3 & 43.2 \\
 & V+T & \underline{55.2} & 18.4 & 75.7 & 68.0 & 73.8 & 15.1 & 35.7 & 14.3 & 48.7 & \underline{72.8} & 74.8 & 14.0 & 12.9 & 44.6 \\
\midrule
\multirow{3}{*}{Adapter} & Vision & 52.9 & \underline{47.8} & 72.2 & 60.1 & 82.0 & \underline{73.6} & \underline{78.6} & 53.7 & 46.8 & 70.8 & 64.4 & 16.7 & \underline{29.2} & 57.6 \\
 & Text & 52.8 & 46.8 & 74.8 & 70.0 & 78.4 & 47.0 & 54.3 & 24.1 & 46.1 & 68.0 & 72.0 & 13.9 & 20.2 & 51.4 \\
 & V+T & 52.5 & 47.2 & 73.3 & 60.8 & \underline{82.8} & 74.3 & 78.3 & \underline{56.1} & \underline{49.6} & 72.4 & 68.4 & 16.9 & 28.7 & \underline{58.6} \\
\midrule
\multirow{3}{*}{Resampler} & Vision & 13.6 & 38.1 & 30.8 & 7.0 & 18.0 & 65.4 & 72.5 & 18.7 & 3.6 & 60.8 & 51.2 & 17.4 & 20.9 & 32.2 \\
 & Text & 13.2 & 38.0 & 31.6 & 6.9 & 17.2 & 65.1 & 72.3 & 20.4 & 5.4 & 51.6 & 48.8 & \underline{18.3} & 20.4 & 31.5 \\
 & V+T & 13.2 & 37.0 & 29.8 & 6.4 & 18.4 & 65.3 & 71.8 & 18.9 & 4.7 & 56.4 & 48.0 & 16.6 & 20.9 & 31.3 \\
\midrule
\rowcolor{gray!15} \textbf{DiVA-Former} & \textbf{V+T} & \textbf{60.9} & \textbf{50.4} & \textbf{78.0} & \textbf{70.5} & \textbf{92.0} & \textbf{79.1} & \textbf{85.3} & \textbf{76.2} & \textbf{71.6} & \textbf{79.1} & \underline{75.6} & \textbf{18.7} & \textbf{35.5} & \textbf{67.1} \\
\bottomrule
\end{tabular}}
\caption{\textbf{Main results on 13 table benchmarks.} The best results are highlighted in \textbf{bold}, and the second-best results are \underline{underlined}. Our model delivers the strongest performance on 12 of the 13 benchmarks and surpasses all baselines in overall average score.}
\label{tab:main_results}
\end{table*}

\section{Experiments}
\label{sec:experiments}

\subsection{Experimental Setup}
To comprehensively evaluate the effectiveness of DiVA-Former, we conduct extensive experiments across 13 public tabular datasets. These datasets encompass four distinct task categories: (1) \textbf{Table Question Answering}, which includes Wiki Table Questions (WTQ) \cite{pasupat-liang-2015-compositional}, Free-form Table Question Answering (FeTaQA) \cite{Nan2021FeTaQAFT}, Tabular And Textual dataset for Question Answering (TAT-QA) \cite{zhu2021tat}, Hierarchical table dataset for question answering (HiTab) \cite{cheng2021hitab}, and Tabular Math Word Problems (TabMWP) \cite{lu2023dynamic}; (2) \textbf{Table Structure Understanding}, comprising Table Size Recognition (TSR), Table Cell Extraction (TCE), Row/Column Extraction (RCE), and Table Cell Retrieval (TCR); (3) \textbf{Table Fact Verification}, featuring TabFact and InfoTabS \cite{2019TabFactA, gupta-etal-2020-infotabs}; and (4) \textbf{Table-to-Text Generation}, covering RotoWire and WikiBio \cite{wiseman-etal-2017-challenges, lebret-etal-2016-neural}. 

We utilize all datasets from a unified public benchmark collection\footnote{\url{https://huggingface.co/datasets/SpursgoZmy/IFT-Data-For-Tabular-Tasks}}, strictly adhering to their original train and test splits. Detailed statistics and descriptions of these datasets are provided in Appendix~\ref{app:datasets}. During evaluation, we maintain the prompt designs from the original works to ensure a strictly fair comparison. We report Accuracy for WTQ, HiTab, TabMWP, TCR, TabFact, and InfoTabS, while for the remaining tasks, we compute the ROUGE-L (F1) score against the standard ground-truth answers. All experiments are conducted using \texttt{Qwen3-VL-8B-Instruct}\footnote{\url{https://huggingface.co/Qwen/Qwen3-VL-8B-Instruct}} as the backbone vision-language model, and the main results are averaged over three independent runs to ensure robustness and reliability.

\subsection{Baselines}
We compare our method against three categories of baselines: \textbf{Direct}, \textbf{Adapter}, and \textbf{Resampler}.
Each baseline category is evaluated under three modality configurations: Text-only, Vision-only, and Vision+Text (V+T). 
For the V+T configuration, we adopt the better-performing vision-first order. The reverse order results are detailed in Appendix~\ref{app:modality_order}.

\paragraph{(1) Direct Input.} 
This baseline directly feeds the tabular input into the LLM without introducing any additional trainable module before the backbone.

\paragraph{(2) Adapter.}
This approach adopts a token-wise two-layer MLP as a lightweight adapter to project input tokens into a representation space better aligned with the LLM \cite{houlsby2019parameterefficienttransferlearningnlp}. 

\paragraph{(3) Resampler.} 
This baseline employs a set of learnable query tokens to extract context, producing a fixed-length sequence of digest tokens. This architecture is analogous to IC-Former \cite{wang2024context} for text inputs and to the Q-Former from BLIP-2 \cite{li2023blip} for visual inputs.

For a fair comparison, all Adapter and Resampler baselines are trained from scratch on the same tabular training data as DiVA-Former.

\begin{figure*}[t]
    \centering
    \includegraphics[width=\textwidth]{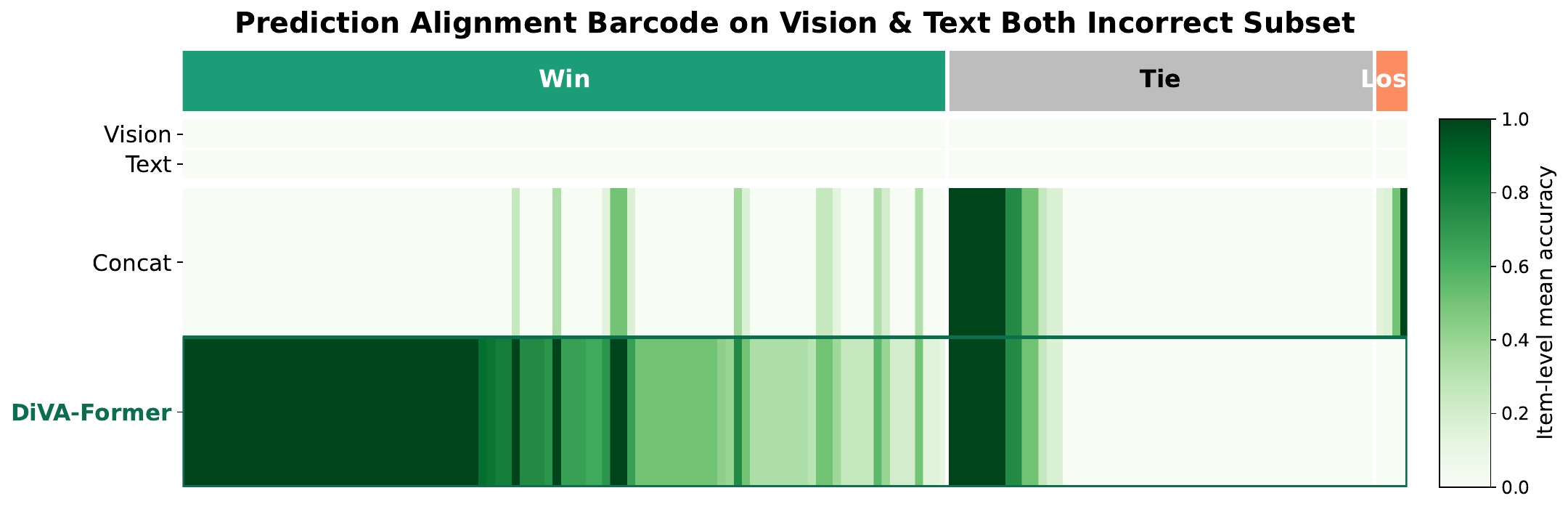}
\caption{Prediction alignment barcode for the subset of examples where both unimodal settings fail. Each column represents a single example, and the color intensity indicates the number of correctly localized target coordinates. We categorize the comparison between DiVA-Former and the direct concat baseline into Win, Tie, and Loss groups. DiVA-Former successfully recovers substantially more of these difficult examples than direct concat, demonstrating that its performance gains stem from leveraging complementary cross-modal information to solve cases that unimodal approaches cannot.}
    \label{fig:barcode}
\end{figure*}

\subsection{Main Results}
\label{sec:mainresults}
Table~\ref{tab:main_results} presents the results of three categories of baselines and our DiVA-Former on 13 datasets. Across this comprehensive benchmark suite, we highlight the following three main findings: 

\begin{itemize}
[nosep,labelwidth=*,leftmargin=1.2em,align=left]
    \item \textbf{Text and vision are highly complementary for 2D structure understanding.}
\end{itemize}
Overall, \emph{vision} inputs tend to outperform \emph{text} inputs on structure understanding benchmarks (TCE), highlighting the benefit of vision tokens in encoding spatial priors. Meanwhile, \emph{text} inputs remain competitive on content-heavy datasets (HiTab), where precise access to symbols and numerical values is critical. Under the Direct Input setting, integrating both modalities typically results in performance on par with the stronger unimodal variant.

\begin{itemize}
[nosep,labelwidth=*,leftmargin=1.2em,align=left]
    \item \textbf{Existing baseline families all exhibit limitations in multimodal fusion.}
\end{itemize}
The Direct Input method rarely leverages the complementary information from both modalities to solve examples that all unimodal variants fail on, as we will show in Section~\ref{sec:gains_from_fusion}. Moreover, the ultra-long context in ``Vision+Text'' weakens the model's ability to retrieve and reason over key information, leading to positional bias, performance degradation, and erroneous downstream predictions~\cite{liu-etal-2024-lost, ICLR2025_a264726e}. As a result, simple direct concatenation cannot consistently outperform unimodal methods across benchmarks.

Adapters performs best among the baselines, especially under the ``Vision+Text'' and ``Vision'' settings. However, ``Vision+Text'' does not significantly outperform ``Vision'', indicating that this method still fails to truly fuse multimodal information. Nevertheless, this observation suggests that continuous visual tokens may be better suited for optimization as trainable ``soft'' embeddings.

The limitations of Resampler are even more pronounced. \emph{First}, its fixed-length learnable queries introduce a strong semantic bottleneck~\cite{yao2024decodecouplingtokencompression, zhong2024enhancingmultimodallargelanguage, hu2025tokenflexunifiedvlmtraining}, which becomes even more problematic when the model must process both modalities. Our experiments show that Resampler is the only method for which the ``Vision+Text'' variant performs \emph{worse} than its unimodal counterparts. We conduct a search over the number of fixed queries and find that the best results are achieved with a fixed number of $K=128$ learnable queries, which is also the setting adopted in prior work~\cite{wang2024context}. Increasing the number of learnable queries does not improve performance, suggesting that learning such queries from scratch cannot reliably extract the necessary information.
\emph{Second}, during training, Resampler tends to hallucinate and overlook fine-grained details, ultimately learning to mimic only the style of the answer~\cite{li2024tokenpackerefficientvisualprojector, Shang_2025}. In our experiments, on datasets that require extracting a precise final answer (WTQ, HiTab, TabMWP and TCR), this method often generates responses in the correct format but with completely incorrect answers, resulting in very low accuracy.

\begin{itemize}
[nosep,labelwidth=*,leftmargin=1.2em,align=left]
    \item \textbf{DiVA-Former addresses these limitations and achieves consistent gains.}
\end{itemize}
Inspired by the observations above, DiVA-Former initializes learnable queries with visual tokens. Continuous visual tokens are well suited for further optimization and for fusing information from new modalities. In this way, the queries no longer start from sample-independent ``empty slots'', but instead begin as visual anchors already located on the feature manifold of the current image. In contrast, fixed learnable queries are prone to a pathological behavior in which multiple queries collapse into redundant and highly similar ``semantic slots''~\cite{yao2024decodecouplingtokencompression}. Our visual-token initialization enables more effective learning of critical information.

In addition, the gating mechanism inside the resampler alleviates the hallucination and fine-grained detail loss caused by training. Starting from strong visual-token initialization, the model can selectively determine whether the text contains useful fine-grained information or complementary evidence for cross-validation, rather than merely learning how to produce fluent answers.

As a result, DiVA-Former substantially outperforms the strongest baselines across nearly all datasets, reaching performance levels that unimodal approaches cannot achieve.

\begin{table}[t]
\centering
\begin{tabular}{lcc}
\toprule
\textbf{Dataset} & \textbf{Direct concat} & \textbf{DiVA-Former} \\
\midrule
TCR      & 15.97\% & 55.23\% {\small \textcolor{green!50!black}{$(\uparrow$39.26\%)}} \\
HiTab    & 4.35\%  & 26.09\% {\small \textcolor{green!50!black}{$(\uparrow$21.74\%)}} \\
TabMWP   & 3.70\%  & 92.59\% {\small \textcolor{green!50!black}{$(\uparrow$88.89\%)}} \\
WTQ      & 5.26\%  & 21.05\% {\small \textcolor{green!50!black}{$(\uparrow$15.79\%)}} \\
TabFact  & 5.56\%  & 42.59\% {\small \textcolor{green!50!black}{$(\uparrow$37.03\%)}} \\
InfoTabS & 5.66\%  & 41.51\% {\small \textcolor{green!50!black}{$(\uparrow$35.85\%)}} \\
\midrule
\textbf{Average} & \textbf{6.75\%} & \textbf{46.51\%} {\small \textcolor{green!50!black}{\textbf{$(\uparrow$39.76\%)}}} \\
\bottomrule
\end{tabular}%
\caption{Recovery rate on examples where both text-only and vision-only inputs fail.}
\label{tab:ratio_comparison}
\end{table}

\subsection{Where Do the Gains Come From?}
\label{sec:gains_from_fusion}

To substantiate our claim that DiVA-Former effectively integrates complementary information from both modalities, we analyze its performance on examples where both unimodal settings fail (Figure~\ref{fig:barcode} and Table~\ref{tab:ratio_comparison}). If a multimodal approach merely selects the better modality for a given task, it should still fail when both text-only and vision-only inputs are incorrect. Conversely, successfully answering these ``both-wrong'' cases provides direct evidence of true multimodal synergy. 

\begin{table*}[t]
\centering
\small
\setlength{\tabcolsep}{3.5pt}
\resizebox{\textwidth}{!}{%
\begin{tabular}{lllccccc cccc cc cc c}
\toprule
\multirow{2}{*}{Gate} & \multirow{2}{*}{Query} & \multirow{2}{*}{Context}
& \multicolumn{5}{c}{Question Answering}
& \multicolumn{4}{c}{Structure Understanding}
& \multicolumn{2}{c}{Fact Verification}
& \multicolumn{2}{c}{Text Generation}
& \multirow{2}{*}{Avg} \\
\cmidrule(lr){4-8}\cmidrule(lr){9-12}\cmidrule(lr){13-14}\cmidrule(lr){15-16}
& & & WTQ & FeTaQA & TAT & HiTab & TabMWP
& TSR & TCE & RCE & TCR
& TabFact & InfoTabS
& RotoWire & WikiBio
& \\
\midrule

$\checkmark$   & Text & Text   & 22.8 & 44.3 & 72.8 & 56.1  & \underline{91.2} & 71.0 & 75.4 & 38.9 & 34.5 & 62.0 & 70.8 & 17.0 & \underline{35.2} & 53.2 \\
$\checkmark$ & Vision  & Vision & 48.6 & \underline{47.9} & 72.8 & 58.8 & 89.2 & \underline{76.7} & 78.5 & 61.3 & \underline{64.2} & 71.2 & \underline{72.4} & \underline{18.3} & 32.3 & 60.9 \\
$\checkmark$ & Text  & Vision   & \underline{60.0} & 45.8 & \underline{74.5} & \underline{65.3} & 91.0 & 75.0 & \underline{85.2} & \underline{73.9} & 63.1 & \underline{76.3} & 69.1 & 17.5 & 34.3 & \underline{63.9} \\
              & Vision   & Text  & 15.7 & 38.6 & 29.8 & 6.6  & 20.8 & 69.5 & 72.0 & 25.4 & 61.2 & 48.4 & 50.4 & 17.8 & 20.3 & 36.7 \\
\midrule
\rowcolor{gray!15}
$\checkmark$  & \textbf{Vision} & \textbf{Text}   & \textbf{60.9} & \textbf{50.4} & \textbf{78.0} & \textbf{70.5} & \textbf{92.0} & \textbf{79.1} & \textbf{85.3} & \textbf{76.2} & \textbf{71.6} & \textbf{79.1} & \textbf{75.6} & \textbf{18.7} & \textbf{35.5} & \textbf{67.1} \\
\bottomrule
\end{tabular}
}
\caption{\textbf{Ablation on modality roles and identity-preserving gates.} ``Query'' indicates the initialization source of the resampler queries, and ``Context'' denotes the token sequence used as Keys/Values in cross-attention. The full DiVA-Former setting (Text context + Vision queries) with identity-preserving gates achieves the best performance.}
\label{tab:ablation}
\end{table*}

\paragraph{Visualizing analysis on TCR.} 
We begin with TCR, which is less influenced by language-generation priors discussed in Section~\ref{sec:related}. As a result, the source of performance gains is more directly observable.
Figure~\ref{fig:barcode} visualizes a subset of examples where both text-only and vision-only inputs fail entirely. 
Each column corresponds to one such example, and darker colors indicate a higher proportion of correctly localized coordinates.

The pattern is striking: even when both unimodal inputs produce no correct predictions, DiVA-Former still partially or fully solves many of these difficult examples, whereas direct concat rarely does so. This suggests that the improvement does not come from simply selecting the better unimodal input. Instead, DiVA-Former effectively extracts and combines complementary signals across modalities to produce accurate predictions that neither branch can generate in isolation. Additional prediction alignment barcode visualizations on the full dataset are provided in Appendix~\ref{appendix:fusion_analysis}.

\paragraph{Quantitative analysis on more benchmarks.} 
To quantify this synergistic effect, we extend the analysis to benchmarks with well-defined example-level correctness, including TCR, HiTab, TabMWP, WTQ, TabFact, and InfoTabS. As before, we isolate the subset of examples for which both unimodal inputs are incorrect and measure the proportion of these cases that are recovered by direct concat and DiVA-Former (Table~\ref{tab:ratio_comparison}). For TCR, evaluation is conducted at the fine-grained coordinate level, where each coordinate is treated as a binary correct/incorrect outcome. The remaining datasets use binary instance-level correctness.

As shown in Table~\ref{tab:ratio_comparison}, DiVA-Former consistently and substantially outperforms direct concat in recovering these difficult both-wrong cases, while direct concat achieves low recovery rates on most datasets, indicating limited ability to make effective use of the concatenated multimodal context. DiVA-Former shows an especially large advantage on TabMWP. On this widely adopted large-scale benchmark, it recovers 92.59\% of such cases.

The gap is particularly pronounced on structure-sensitive tasks such as TCR, TabFact, and InfoTabS. The gain on WTQ is smaller, which is consistent with WTQ being a highly text-centric table QA benchmark. Although direct concat appears to recover a nontrivial fraction of TCR cases (15.97\%), this result should be interpreted with caution, as it is largely an artifact of TCR's coordinate-level evaluation, which makes partial recovery easier to observe than strict exact-match metrics. 

In summary, both the qualitative heatmap
and the quantitative recovery analysis
demonstrate that DiVA-Former succeeds by integrating complementary multimodal evidence, enabling the LLM to solve cases intractable for either modality alone.

\begin{figure*}[t]
\centering
\includegraphics[width=0.33\textwidth]{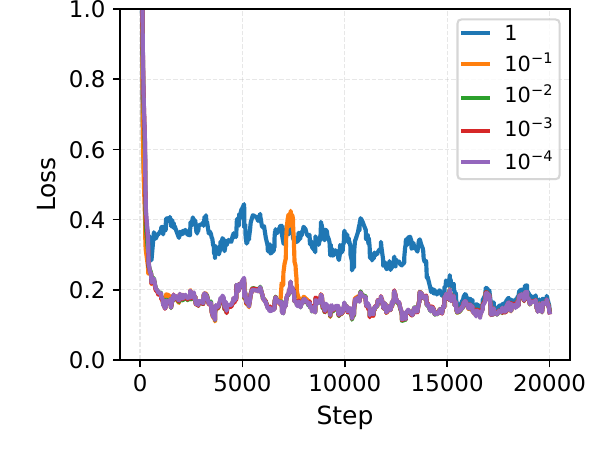}\hfill
\includegraphics[width=0.33\textwidth]{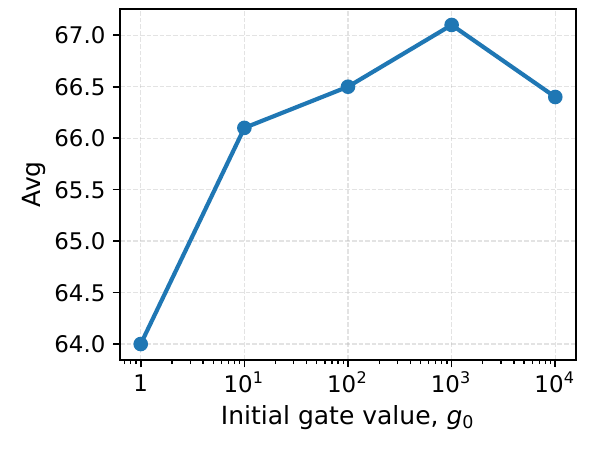}\hfill
\includegraphics[width=0.33\textwidth]{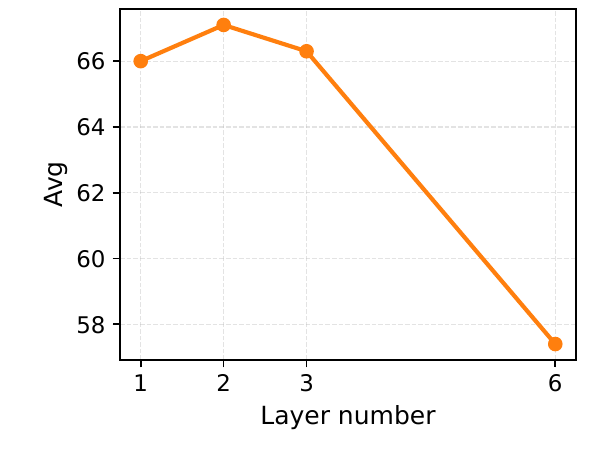}

\vspace{-10pt} 

\caption{\textbf{Hyperparameter analysis of DiVA-Former.} Left: training loss curves under different initial gate values $g_0$. Middle: final average score across the 13 benchmarks under different initial gate values $g_0$. Right: final average score under different numbers of DiVA-Former layers.}
\label{fig:hyperparam}
\end{figure*}

\subsection{Ablation Studies}
\label{sec:ablation}

Having established that the gains of DiVA-Former come from genuine multimodal fusion, we next examine how different architectural choices affect performance in Table~\ref{tab:ablation}.

\paragraph{Role assignment for different modalities.}
For the query side, the results are consistent with our earlier observations: initializing the queries from \emph{vision} tokens leads to clearly better overall performance. By contrast, using \emph{text} embeddings as queries causes a substantial degradation, suggesting that text-side representations, being tightly anchored to discrete lexical tokens, are less suitable as learnable queries for further optimization.

For the context side, using \emph{vision} for both context and queries already yields promising results as a unimodal approach. However, this configuration still underperforms the full model, particularly on question answering datasets focused on extracting specific numerical answers, where precise cell content, symbols, and numerical details must be preserved faithfully. This indicates that using \emph{text} as context is necessary to provide complementary information from the additional modality.

Using \emph{text} as queries and \emph{vision} as context can be regarded as a modality-swapped variant of our DiVA-Former, with the roles of query and context reversed. Although this setup benefits from additional cross-modal information and achieves reasonably good results, as noted earlier, text representations are tightly bound to discrete lexical tokens and are therefore ill-suited to serve as learnable queries. More importantly, this design departs from the original motivation of the query resampler, which is to distill complex inputs into a compact set of equivalent representations. Here, a relatively long set of learnable queries is instead used to extract comparatively short visual information, resulting in an inefficient scheme. It also causes the computational cost of the downstream LLM to grow rapidly with the number of digest vectors.

Overall, using \emph{vision} as queries and \emph{text} as context is not only superior in terms of empirical results and efficiency, but also a highly principled design choice. As the preceding analysis indicates, this configuration is both elegant and theoretically grounded. By aligning each modality with its inherent strengths, it facilitates robust structural abstraction on the query side while maintaining high-fidelity textual precision on the context side.

\paragraph{Identity-preserving gates are essential.}
Removing the gates causes a dramatic drop in performance, even under the same modality assignment. 
The results obtained without the gating mechanism closely mirror those of the query resampler discussed in Section~\ref{sec:mainresults}. It often generates responses in the correct format but with completely incorrect answers, resulting in very low accuracy on datasets that require extracting a precise final answer.
This suggests that the identity-preserving gates play a critical role in stabilizing optimization, preventing DiVA-Former from drifting too far away from the original visual priors.

\subsection{Hyperparameter Analysis}
\label{sec:hyperparam}

We next investigate two key hyperparameters of DiVA-Former: the gate initialization value $g_0$ and the number of resampler layers.

\paragraph{Gate initialization is crucial for stable fusion.}
Although the gates are learnable and can be adjusted during training, their initialization has a strong impact on early-stage optimization.
As shown in Figure~\ref{fig:hyperparam} (Left), setting $g_0 = 1$ leads to consistently higher training loss, indicating unstable updates when the residual branch is fully activated from the beginning.
A moderately small initialization ($10^{-1}$) can also be unstable, exhibiting a sudden spike in loss during training.
In contrast, smaller initialization values converge quickly and remain stable, which is consistent with the intended near-identity behavior: preserving visual priors while gradually incorporating textual evidence.
Figure~\ref{fig:hyperparam} (Middle) further shows that smaller $g_0$ values yield better final accuracy on average.
We adopt $g_0 = 10^{-3}$ as the default setting.


\paragraph{A shallow design is sufficient.}
Figure~\ref{fig:hyperparam} (right) studies the impact of resampler depth.
We find that configurations with 1 to 3 layers yield comparable results, while increasing the depth to 6 leads to a marked performance drop. 
This confirms that, with strong visual-token initialization and identity-preserving gates, a shallow design adequately captures the required transformations. 
Notably, the robust performance of a single-layer configuration further suggests that effective vision-text fusion in our model stems from architectural efficiency rather than brute-force parameter scaling. 
Consequently, we adopt $N = 2$ as the default setting.

\section{Conclusions}
\label{sec:conclusions}
In this paper, we demonstrate that visual and textual modalities provide complementary information for 2D structure understanding. Motivated by this observation, we propose DiVA-Former, a lightweight fusion architecture that uses visual tokens as anchors to digest lengthy textual context,
thereby achieving an effective complementary fusion of both modalities.
Across 13 benchmarks, DiVA-Former surpasses the upper bound achievable by unimodal information alone.

\section*{Limitations}
\label{sec:limitations}
While our results are promising for multimodal table understanding by using complementary vision-text information, several limitations remain.

\begin{itemize}[leftmargin=*]

\item First, although our evaluation covers 13 datasets across four task categories, all of them provide predefined training and test splits. Therefore, we have not yet examined whether our method can generalize equally well to table understanding tasks without task-specific training data. 

\item Second, our experiments are conducted using only a recent model, \texttt{Qwen3-VL-8B-Instruct}, as the backbone. Although this setting is sufficient to validate the effectiveness of DiVA-Former, further experiments on a broader range of multimodal LLMs would provide a more comprehensive understanding of its generalizability.
\end{itemize}

\bibliography{custom}

\appendix
\clearpage

\section{The Bottleneck of Visual Text Compression Methods}
\label{app:soft}

This section provides empirical evidence for our claim that existing context compression methods still struggle to surpass uncompressed baselines. In particular, we focus on the limitations of relying on the visual modality alone to read textual content.

We revisit the current state-of-the-art visual text compression framework~\cite{cheng2025glyphscalingcontextwindows}. Their method renders long text into images and processes the rendered inputs with a vision-language model. The framework consists of three main stages: continual pre-training on rendered long-text data, LLM-driven genetic search for optimal rendering configurations, and post-training with supervised fine-tuning and reinforcement learning. The original paper reports substantial token compression while maintaining accuracy competitive with strong long-context LLMs such as Qwen3-8B across a range of benchmarks.

In Table~\ref{tab:soft} our reproduction largely matches the performance reported in the original paper, confirming the effectiveness of the \emph{Digesting Context into Soft Embeddings} paradigm. However, when we directly feed the same content as text instead of rendered images, we observe a consistent trend: direct text input still outperforms visual compression on most benchmarks, despite the extensive optimization applied to the rendering-based pipeline. Moreover, providing both visual and textual inputs together does not yield further gains.

More importantly, this comparison highlights the key distinction of our approach. Although our method also reduces the number of tokens passed to the target LLM, it does not rely on visual token alone. Instead, it leverages complementary vision-text information to capture 2D structural cues that are difficult to preserve in purely textual or purely visual compression pipelines. This is especially important for structured inputs such as tables, where spatial layout is a critical part of the semantics. As a result, our method overcomes the bottleneck of visual text compression approaches, surpasses the fully uncompressed text-input baseline, and demonstrates the value of using visual tokens for better understanding rather than merely as a compression medium \cite{zhang-etal-2024-modeling, wang-etal-2024-docllm}.

\begin{table}[t]
\centering
\resizebox{\columnwidth}{!}{
\begin{tabular}{lrrr}
\toprule
Modality & Vision & Text & V + T \\
\midrule
2wikimqa & 74.12 & \textbf{78.26} & 77.43 \\
dureader & 33.54 & 40.49 & \textbf{41.51} \\
gov\_report & 25.47 & \textbf{26.26} & 25.79 \\
hotpotqa & 62.86 & \textbf{74.33} & 74.30 \\
multi\_news & 21.62 & \textbf{23.06} & 22.69 \\
multifieldqa\_en & 54.66 & \textbf{59.55} & 58.42 \\
multifieldqa\_zh & 42.25 & \textbf{64.34} & 62.92 \\
musique & 66.13 & \textbf{68.69} & 68.19 \\
narrativeqa & 49.97 & \textbf{53.04} & 52.95 \\
passage\_retrieval\_en & 98.19 & \textbf{100.00} & 100.00 \\
passage\_retrieval\_zh & 96.20 & 99.75 & \textbf{100.00} \\
qasper & 63.47 & \textbf{65.90 }& 64.82 \\
qmsum & \textbf{77.61} & 77.17 & 76.00 \\
vcsum & \textbf{0.14} & 0.03 & 0.02 \\
repobench-p & 57.63 & \textbf{64.90} & 61.09 \\
\midrule
Avg & 54.92 & \textbf{59.72} & 59.08 \\
\bottomrule
\end{tabular}}
\caption{Comparison between visual compression and direct text input on long-context benchmarks. Despite strong performance from the visual compression framework, direct text input remains better on most datasets, indicating that current soft-embedding methods still incur non-trivial information loss.}
\label{tab:soft}
\end{table}

\section{Datasets}
\label{app:datasets}

This work evaluates models on a diverse collection of table-centric benchmarks, covering
(i) table question answering (QA), (ii) table-based fact verification, (iii) table-to-text generation,
and (iv) basic table-structure understanding.

\subsection{Table Question Answering}

\paragraph{Wiki Table Questions (WTQ).}
WTQ is a classic benchmark for semantic parsing on semi-structured Wikipedia tables, where crowd workers
ask questions and provide answers grounded in a single table. The questions often require non-trivial table
operations.

\paragraph{FeTaQA.}
FeTaQA targets generative table QA. Instead of short-span answers, the expected output is a free-form,
variable-length natural language response grounded in evidence tables cells.

\paragraph{TAT-QA.}
TAT-QA focuses on hybrid contexts consisting of a table plus accompanying textual descriptions from real-world
financial reports. Many questions require multi-hop reasoning and numerical computation; the original release
includes answer scales and derivations for numerically grounded questions.

\paragraph{HiTab.}
HiTab contains \emph{hierarchical} tables and supports both table QA and table-to-text style generation.
Hierarchical indexing and implicit aggregation relations make reasoning substantially more challenging than
flat-table settings.

\paragraph{Tabular Math Word Problems (TabMWP).}
TabMWP contains grade-level math word problems aligned with tabular contexts. Each problem is annotated with
gold solutions to expose multi-step reasoning, and is designed to probe mathematical reasoning over
heterogeneous (text + table) inputs.

\subsection{Table-based Fact Verification}

\paragraph{TabFact.}
TabFact provides Wikipedia tables as evidence for verifying human-written statements, labeled as entailed or
refuted. The dataset is large-scale and includes statements that require both linguistic and symbolic reasoning.

\paragraph{InfoTabS.}
InfoTabS formulates table understanding as natural language inference (NLI) over Wikipedia infobox tables, with
three-way labels (entailment/contradiction/neutral).

\subsection{Table-to-Text Generation}

\paragraph{RotoWire.}
RotoWire is a data-to-document benchmark that pairs NBA box-score statistics with human-written game summaries,
testing the ability to generate faithful, coherent multi-sentence descriptions from structured records.

\paragraph{WikiBio.}
WikiBio contains Wikipedia infoboxes paired with corresponding biography text, enabling evaluation of
structure-to-text generation (biography generation) from attribute-value records.

\subsection{Table Structure Understanding}

Recent studies suggest that strong performance on downstream table tasks does not necessarily imply that a model truly understands table structure \cite{10.1145/3616855.3635752}. To ensure a comprehensive evaluation, we adopt the four basic table-structure tasks:
(i) \textbf{Table Size Recognition (TSR)} for predicting the number of rows/columns,
(ii) \textbf{Table Cell Extraction (TCE)} for extracting the content at a given row/column index,
(iii) \textbf{Row/Column Extraction (RCE)} for retrieving all cells in a specified row (or column).
These tasks directly evaluate whether a model can reliably follow table coordinates and structure, and
(iv) \textbf{Table Cell Retrieval (TCR)} for recovering the row/column index given a cell value.

\subsection{Prompt}


\begin{promptbox}[blue]{WTQ Prompt}
Observe the illustrated table involved with 1931 Tour de France and give a brief response to address the subsequent question:\\
did each winner win on stage with mountain(s)?\\
Format your final answer as a JSON, using the structure \{"answer": [<a list of answer strings>]\}.
\end{promptbox}

\begin{promptbox}[green]{TAT-QA Prompt}
Relevant paragraphs: ...\\
Given the table and relevant paragraphs, answer the following question briefly. Present the final answer in a JSON format \{"answer": [<a list of answer strings>]\}.\\
What is the total guarantee in respect of Section 106 planning obligation liabilities at Barton Square which at 31 December 2018?
\end{promptbox}

\begin{promptbox}[orange]{HiTab Prompt}
This excel table is related to 'citizenship rates among immigrants aged 18 and older who arrived in canada five to nine years ago, by selected socio-demographic characteristics, 1991 to 2016', answer the following question concisely. Show your answer in the JSON format \{"answer": [<a list of answer strings>]\}.\\
between 1996 and 2016, among which group of people was the decline greater, among those with poorer official language skills or among those whose mother tongue was english or french?
\end{promptbox}

\begin{promptbox}[purple]{TabMWP Prompt}
Please solve the problem based on the given table about 'Donations'. In the end, output your final answer using the JSON format: \{"answer": "<YOUR ANSWER>"\}. \\
Problem: A philanthropic organization ... How much did Tamir donate to clean water?\\
Your Solution:
\end{promptbox}


\begin{promptbox}[red]{FeTaQA Prompt}
Based on this table about 'Pale Waves' and its title is 'Music videos', provide the answer to the following question: What year was Pale Waves' music video for "Television Romance" released and who directed it?
\end{promptbox}

\begin{promptbox}[brown]{TSR Prompt}
This is a table. Can you figure out the row and column numbers for this particular table? The final result should be presented in the JSON format of \{"row\_number": "m", "column\_number": "n"\}.
\end{promptbox}

\begin{promptbox}[magenta]{TCE Prompt}
What are the cells' value in the following positions in the table? Represent each cell value in the JSON format \{"row\_id":"m", "column\_id":"n", "cell\_value":"<Corresponding Cell Value>"\}. For instance, \{"row\_id":"5", "column\_id":"4", "cell\_value":"missouri 4"\}.\\
row 2 and column 7\\
row 6 and column 5\\
row 7 and column 4
\end{promptbox}

\begin{promptbox}[cyan]{RCE Prompt}
Give you a column index and a table, extract the corresponding cell values of this column:\\
column 5\\
The cells in a column should be presented in the JSON format of \{"column\_id":"<column index>", "cell\_list":"<a list of cells in this column>"\}, e.g., \{"column\_id":"3", "cell\_list": ["xxx", "yyy", "zzz"]\}.
\end{promptbox}

\begin{promptbox}[olive]{TCR Prompt}
Analyze the provided table and identify the locations of these cells:\\
1. '24.04.1949';\\
2. '07.07.1982';\\
3. 'Grenoble';\\
Provide the cell locations using the JSON format \{'value': '<cell value>', 'location': (Row\_ID, Column\_ID)\}, where row and column IDs start from 1. Denote the location as 'DOES NOT EXIST' if a cell does not exist in the table.
\end{promptbox}


\begin{promptbox}[pink]{InfoTabS Prompt}
Review the table as a premise and decide if it upholds or contradicts the above claim. Classify the claim as 'neutral' if the table information is not enough for a final conclusion. Format your final answer as a JSON, using the structure \{"answer": "<YOUR ANSWER>"\}.
\end{promptbox}

\begin{promptbox}[violet]{TabFact Prompt}
This is a table named 'list of intel pentium dual - core microprocessors', please determine whether the given sentence is substantiated or opposed by the table information. Your final answer should be in the JSON structure, formatted as \{"answer": "<YOUR ANSWER>"\}.\\
the processor with a part number of lf80537 ge0251 mn has a fsb of 533 mt / s
\end{promptbox}


\begin{promptbox}[darkgray]{RotoWire Prompt}
Examine the data records in this, which depict an NBA game held on 2016-11-26. Use these data statistics to generate a comprehensive summary of the game ...
\end{promptbox}

\begin{promptbox}[teal]{WikiBio Prompt}
This shows a fact table describing a person named 'Mohd norhafiz zamani misbah', write a concise biography of this person based on the provided information.
\end{promptbox}

\newcommand{\redcell}[1]{\textcolor{red}{#1}}

\begin{table*}[t]
\centering
\resizebox{\textwidth}{!}{
\begin{tabular}{llcccccccccccccc}
\toprule
\multirow{2}{*}{Method} & \multirow{2}{*}{Modality} 
& \multicolumn{5}{c}{Question Answering} 
& \multicolumn{4}{c}{Structure Understanding} 
& \multicolumn{2}{c}{Fact Verification} 
& \multicolumn{2}{c}{Text Generation} 
& \multirow{2}{*}{Overall} \\
\cmidrule(lr){3-7} \cmidrule(lr){8-11} \cmidrule(lr){12-13} \cmidrule(lr){14-15}
& & WTQ & FeTaQA & TAT & HiTab & TabMWP & TSR & TCE & RCE & TCR & TabFact & InfoTabS & RotoWire & WikiBio & \\
\midrule

\multirow{4}{*}{Direct}
& Vision & 55.1 & 17.4 & 75.8 & 63.5 & 74.7 & 14.8 & 36.1 & 13.8 & 39.5 & 72.0 & 77.6 & 13.9 & 13.1 & 43.6 \\
& Text   & 54.6 & 19.2 & 73.9 & 68.4 & 75.3 & 14.2 & 24.2 & 14.8 & 43.0 & 72.4 & 72.8 & 15.2 & 13.3 & 43.2 \\
& V+T    & 55.2 & 18.4 & 75.7 & 68.0 & 73.8 & 15.1 & 35.7 & 14.3 & 48.7 & 72.8 & 74.8 & 14.0 & 12.9 & 44.6 \\
& \redcell{T+V} 
& \redcell{54.8} & \redcell{18.3} & \redcell{70.8} & \redcell{68.5} & \redcell{76.2} 
& \redcell{14.0} & \redcell{35.4} & \redcell{14.5} & \redcell{45.9} 
& \redcell{70.8} & \redcell{75.2} & \redcell{13.5} & \redcell{13.6} & \redcell{44.0} \\
\midrule

\multirow{4}{*}{Adapter}
& Vision & 52.9 & 47.8 & 72.2 & 60.1 & 82.0 & 73.6 & 78.6 & 53.7 & 46.8 & 70.8 & 64.4 & 16.7 & 29.2 & 57.6 \\
& Text   & 52.8 & 46.8 & 74.8 & 70.0 & 78.4 & 47.0 & 54.3 & 24.1 & 46.1 & 68.0 & 72.0 & 13.9 & 20.2 & 51.4 \\
& V+T    & 52.5 & 47.2 & 73.3 & 60.8 & 82.8 & 74.3 & 78.3 & 56.1 & 49.6 & 72.4 & 68.4 & 16.9 & 28.7 & 58.6 \\
& \redcell{T+V} 
& \redcell{51.2} & \redcell{46.8} & \redcell{73.0} & \redcell{60.5} & \redcell{82.3} 
& \redcell{73.9} & \redcell{78.4} & \redcell{55.7} & \redcell{47.8} 
& \redcell{72.3} & \redcell{67.9} & \redcell{17.0} & \redcell{28.6} & \redcell{58.1} \\
\midrule

\multirow{4}{*}{Resampler}
& Vision & 13.6 & 38.1 & 30.8 & 7.0 & 18.0 & 65.4 & 72.5 & 18.7 & 3.6 & 60.8 & 51.2 & 17.4 & 20.9 & 32.2 \\
& Text   & 13.2 & 38.0 & 31.6 & 6.9 & 17.2 & 65.1 & 72.3 & 20.4 & 5.4 & 51.6 & 48.8 & 18.3 & 20.4 & 31.5 \\
& V+T    & 13.2 & 37.0 & 29.8 & 6.4 & 18.4 & 65.3 & 71.8 & 18.9 & 4.7 & 56.4 & 48.0 & 16.6 & 20.9 & 31.3 \\
& \redcell{T+V} 
& \redcell{13.3} & \redcell{37.3} & \redcell{28.5} & \redcell{6.3} & \redcell{17.6} 
& \redcell{65.2} & \redcell{71.9} & \redcell{18.8} & \redcell{4.9} 
& \redcell{55.1} & \redcell{48.6} & \redcell{16.8} & \redcell{20.5} & \redcell{31.1} \\
\midrule

DiVA-Former & V+T & 60.9 & 50.4 & 78.3 & 70.8 & 92.0 & 78.6 & 85.3 & 75.6 & 71.6 & 79.1 & 75.6 & 18.7 & 35.5 & 67.1 \\
\bottomrule
\end{tabular}
}
\caption{Main results with different  modality order. Rows with modality \textcolor{red}{T+V} are highlighted in red.}
\label{tab:order}
\end{table*}

\section{Effect of Modality Order}
\label{app:modality_order}

In addition to the primary multimodal setting reported in the main text, we also evaluate the reverse input order, \textbf{Text+Vision (T+V)}, for all three baseline families: \textbf{Direct Input}, \textbf{Adapter}, and \textbf{Resampler}. The complete results are included in Table~\ref{tab:order}, where the T+V rows are highlighted in red.

Overall, ``Text+Vision'' consistently performs worse than ``Vision+Text'' across all three baseline families. Specifically, the overall score drops from 44.6 to 44.0 for Direct Input, from 58.6 to 58.1 for Adapter, and from 31.3 to 31.1 for Resampler when switching from ``Vision+Text''  to ``Text+Vision''. Although the gap is moderate, the trend is consistent across methods.

We attribute this behavior to the fact that the underlying vision-language model is better aligned with an image-first input format. Prior work has shown that model performance can vary substantially with prompt format and that stronger reliance on visual tokens is often important for robust multimodal grounding \cite{bi-etal-2025-llava, zhuo-etal-2024-prosa, xu-etal-2023-multiinstruct, xu-etal-2025-mitigating}. In our setup, placing visual tokens before textual tokens appears to better align with this behavior, leading to slightly more effective fusion.

\begin{figure*}[t]
    \centering
    \includegraphics[width=\textwidth]{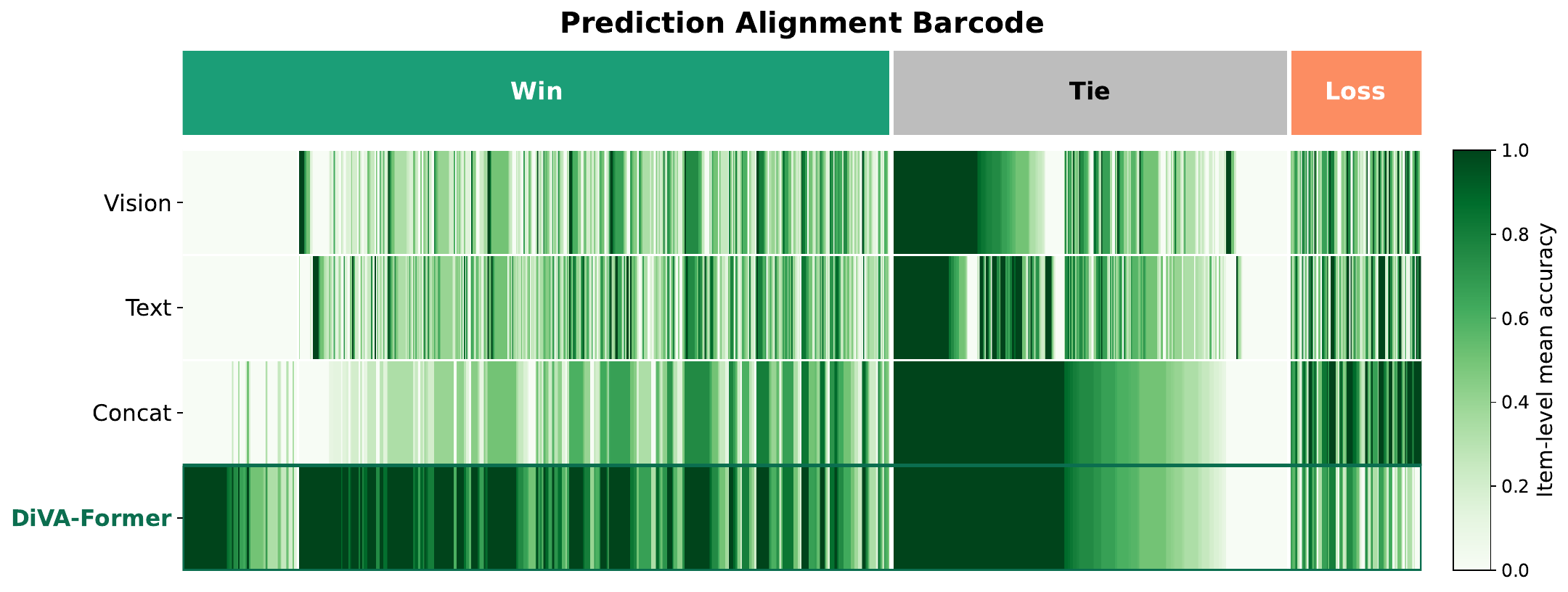}
\caption{\textbf{Full-set per-example prediction alignment analysis on TCR.} The left region corresponds to examples where DiVA-Former outperforms direct concatenation, the middle region indicates ties, and the right region contains examples where DiVA-Former underperforms. Notably, many examples in the left region are cases that unimodal models and direct concatenation fail to solve at all, while even on some examples in the right region DiVA-Former still recovers a substantial number of coordinates.}
    \label{fig:barcode_appendix}
\end{figure*}

\section{Additional Analysis of Fusion Gains on the Full TCR Set}
\label{appendix:fusion_analysis}

We provide here the full-set version of the TCR analysis discussed in Section~\ref{sec:gains_from_fusion}. Unlike the main-text figure, which isolates the hardest cases where both unimodal models fail completely, this appendix includes all evaluation examples.

Figure~\ref{fig:barcode_appendix} presents the per-example alignment heatmap for this complete dataset, where, as in the main text, color intensity indicates localization accuracy. We group these examples into three regions: the left region contains examples where DiVA-Former performs better than direct concatenation, the middle region indicates ties, and the right region contains examples where DiVA-Former performs worse.

Two patterns are particularly clear. First, a large portion of the gains comes from examples that cannot be solved by either unimodal model or by naive ``Vision+Text'' concatenation, yet become partially or fully correct under DiVA-Former. This supports our central claim that effective fusion should create \emph{new} correct predictions, rather than merely preserve unimodal successes. Second, even on examples where DiVA-Former is overall worse than direct concatenation, it often still recovers a non-trivial number of correct coordinates. This suggests that the fusion mechanism improves structural grounding more broadly, rather than only helping on a small set of isolated examples.

Taken together with the harder subset shown in the main text, these full-set results reinforce the same conclusion: the advantage of DiVA-Former is not simply due to better unimodal competence or naive multimodal aggregation, but to more effective cross-modal fusion.

\end{document}